\documentclass[
]{ceurart}

\usepackage{listings}
\usepackage[english]{babel}
\usepackage{amsthm}

\theoremstyle{definition}
\newtheorem{definition}{Definition}[section]

\usepackage{graphicx}
\usepackage{hyperref}
\usepackage{amsmath}
\usepackage{amsfonts}

\usepackage{amsthm}
\theoremstyle{definition}

\usepackage{caption}
\usepackage{subcaption}

\usepackage{multirow}
\usepackage[normalem]{ulem}
\useunder{\uline}{\ul}{}

\begin{document}

\copyrightyear{2022}
\copyrightclause{Copyright for this paper by its authors.
  Use permitted under Creative Commons License Attribution 4.0
  International (CC BY 4.0).}

\conference{OAEI’23: The 18th International Workshop on Ontology Matching, November 7th, 2023, Athens, Greece}

\title{Contextualized Structural Self-supervised Learning for Ontology Matching}


\author{Zhu Wang}[%
orcid=0000-0001-6374-8735,
email=zwang260@uic.edu,
url=https://ellenzhuwang.github.io,
]

\address[1]{ADVIS Lab, Dept of Computer Science\\University of Illinois at Chicago, Chicago IL 60607, USA}


\begin{abstract}
Ontology matching (OM) entails the identification of semantic relationships between concepts within two or more knowledge graphs (KGs) and serves as a critical step in integrating KGs from various sources. Recent advancements in deep OM models have harnessed the power of transformer-based language models and the advantages of knowledge graph embedding. Nevertheless, these OM models still face persistent challenges, such as a lack of reference alignments, runtime latency, and unexplored different graph structures within an end-to-end framework. In this study, we introduce a novel self-supervised learning OM framework with input ontologies, called LaKERMap. This framework capitalizes on the contextual and structural information of concepts by integrating implicit knowledge into transformers. Specifically, we aim to capture multiple structural contexts, encompassing both local and global interactions, by employing distinct training objectives. To assess our methods, we utilize the Bio-ML datasets and tasks. The findings from our innovative approach reveal that LaKERMap surpasses state-of-the-art systems in terms of alignment quality and inference time. Our models and codes are available here: \url{https://github.com/ellenzhuwang/lakermap}. 
\end{abstract}

\maketitle

\section{Introduction}

An \textit{ontology} or \textit{knowledge graph (KG)} provides a vocabulary to describe a domain of knowledge \cite{cruz2005role}. These ontologies encompass millions of concepts within the domain \cite{meilicke-thesis}. The examples of concepts are such as \textit{`angiosarcoma', `sarcoma', `hemangiosarcoma'}, and relations between them, are such as \textit{`subclass of' and `synonym'}. However, many ontologies have been developed independently across different systems and domains, leading to heterogeneity and concept naming ambiguity. To facilitate knowledge sharing and reuse, ontology matching (OM) plays a crucial role in knowledge engineering and semantic data integration.


Ontology matching aims to establish semantic correspondences between concepts from different ontologies~\cite{euzenat2007book}. Traditional OM systems, such as AgreementMakerLight (AML) \cite{faria2013agreementmakerlight} and LogMap \cite{jimenez2011logmap}, have consistently obtained top results in the Ontology Alignment Evaluation Initiative (OAEI) \footnote{\url{http://oaei.ontologymatching.org/}}. These systems mainly apply lexical matching algorithms but fail to capture semantic information between the concepts. For instance, distinguishing between `first lumbrical muscle of the foot' and `second lumbrical muscle of the foot' based on lexical matching presents a challenge, as they are distinct concepts semantically.

Recent pre-trained language models (PLMs), such as BERT\cite{devlin2018bert}, BioBert\cite{alsentzer-etal-2019-publicly} and PubMedBERT \cite{pubmedbert} have achieved dominant performance on various Natural Language Processing (NLP) tasks. Note that, BioBERT and PubMedBERT are tailored for medical domain corpus. Thus, systems like MELT\cite{hertling2021matching} and BERTMap\cite{he2021bertmap} have applied the transformer-based models, demonstrating that the use of these pre-trained models leads to notable improvements in ontology matching tasks. Nevertheless, these systems fine-tune on the synonyms of the concepts, disregarding the abundant graph structures between the concepts.

To address the challenge of learning structural information, knowledge graph embedding-based methods have shown promising results. Examples include MutliOM \cite{li2019multi} and AMD \cite{wang2022amd}, which learn relations from the triplets by employing translation-based algorithms \cite{bordes2013translating}, \cite{lin2015learning}. However, these approaches typically focus solely on learning the relational embeddings, without considering contextual semantics, such as language embeddings of labels/names of the concepts.

Motivated by the aforementioned observations, we propose a novel ontology matching method, LaKERMap (\textbf{La}nguage and \textbf{K}nowledge graph \textbf{E}mbedding \textbf{R}epresentation \textbf{Map}ping). This method employs two transformer encoders to incorporate both contextual and structural information in the ontologies. Specifically, LaKERMap utilizes self-supervised training on designed objectives at both the triplet and path levels, such as triplets contrastive learning and masked concept prediction in paths. During the mapping inference, we predict candidates using the pre-trained model and select the final alignments with a relation regularization based on knowledge graph embedding methods.

To evaluate the effectiveness of LaKERMap, we conduct multiple experiments on the Bio-ML track \cite{he2022machine} in the annual OAEI competitions. When compared to the state-of-the-art systems in OAEI 2022, our experimental results demonstrate that LaKERMap consistently outperforms baseline systems across various evaluation metrics. Moreover, considering inference latency, LaKERMap generates alignments for large ontologies within minutes. In summary, our contributions are as follows: (1) We propose LaKERMap, a novel ontology matching system that infuses both contextual and structural information into transformer models. (2) We introduce multiple training objectives with carefully designed strategies. These objectives are beneficial for learning contextualized structural representations both locally and globally. (3) Extensive evaluations on different datasets reveal that LaKERMap outperforms state-of-the-art ontology baseline systems in terms of runtime, recall, and F-score.




\section{Preliminaries}
\subsection{Transformers}
Many PLMs apply a  multi-layer Transformer architecture to encode texts \cite{devlin2018bert}, and the core component is the multi-head self-attention (MHA) \cite{vaswani2017attention}. Given input $X \in \mathbb{R} ^{N \times d}$ as a sequence of N tokens, the $L$-th layer of MHA is derived as:

\begin{equation} \label{eq:mha}
    f^L = MHA(Q^L,K^L,V^L) = concat (Z_1^L, \ldots , Z_H^L) W^L_0 
\end{equation}

\begin{equation}
    Z_h^L = SA(Q_h^L,K_h^L,V_h^L) = softmax (\frac{Q_h^L (K_h^L)^\top}{\sqrt{d_q}})V_h^L
\end{equation}

\begin{equation}
    Q_h^L = X W_{Q,h}^L, K_h^L = X W_{K,h}^L, V_h^L = X W_{V,h}^L
\end{equation}

where $W_0$ is a linear projection matrix, each head $Z_h$ is a self-attention (SA) module,  $h\in[1,H]$ , and $W_{Q,h},W_{K,h},W_{V,h}$ are query, key, value matrices to be learned by the model.

\subsection{Problem Formulation}\label{sec:problem}

Our general goal is to learn both contextual and structural information from the ontologies to enhance ontology matching. Specifically, we aim to employ self-supervised training on multiple tasks for local and global concept representation learning. Given an ontology $O$, we can extract their concept sets as $C$ and relation sets as $R$. Triplets $\in O $ are defined as $(h,r,t)$, where $h \in C$ is head concept, $t \in C$ is tail concept, and $r \in R$ is the relation between head and tail concept. We utilize the triplets in all the public ontologies as our training set.


\begin{definition}[Ontology Matching]
Ontology matching is finding a set of mappings M between the concepts of source ontology as $O$ and target ontology as $O'$.
\end{definition}
In this work, we mainly focus on the equivalence matching. A mapping is a tuple $(c, c', \equiv, s)$, where $c \in O$ and $c' \in O'$ are concepts, , and $s \in [0,1]$ indicates the probability of concept $c$ and $c'$ are equivalent as follow:

\begin{equation} \label{eq:score}
s(c \equiv c') = \cos (( f(c),  f(c')))
\end{equation}

where $f(\cdot)$ is the feature extraction module, such as the output final layer from Eq \ref{eq:mha}, $\cos $ is the cosine similarity between the feature embeddings of given concepts. 
\section{Proposed method}


In this section, we detail our proposed LaKERMap framework. Specifically, we design multiple training objectives to incorporate structural context at the triplet and path levels. We first introduce how we jointly encode text and relations between concepts using masked triplet contrastive learning and relation prediction for triplets.

Furthermore, global structure information is crucial for learning KG structures, as similar concepts often have similar parents or children. To this end, we aim to distinguish positive and negative paths using a transformer encoder. Moreover, ontologies typically contain multiple, lengthy paths leading from the root to given concepts. Consequently, masked concept prediction at the path level is beneficial for representing the global structure of a KG.

\begin{figure*}[hbt]
\centering
\includegraphics[width=\textwidth]{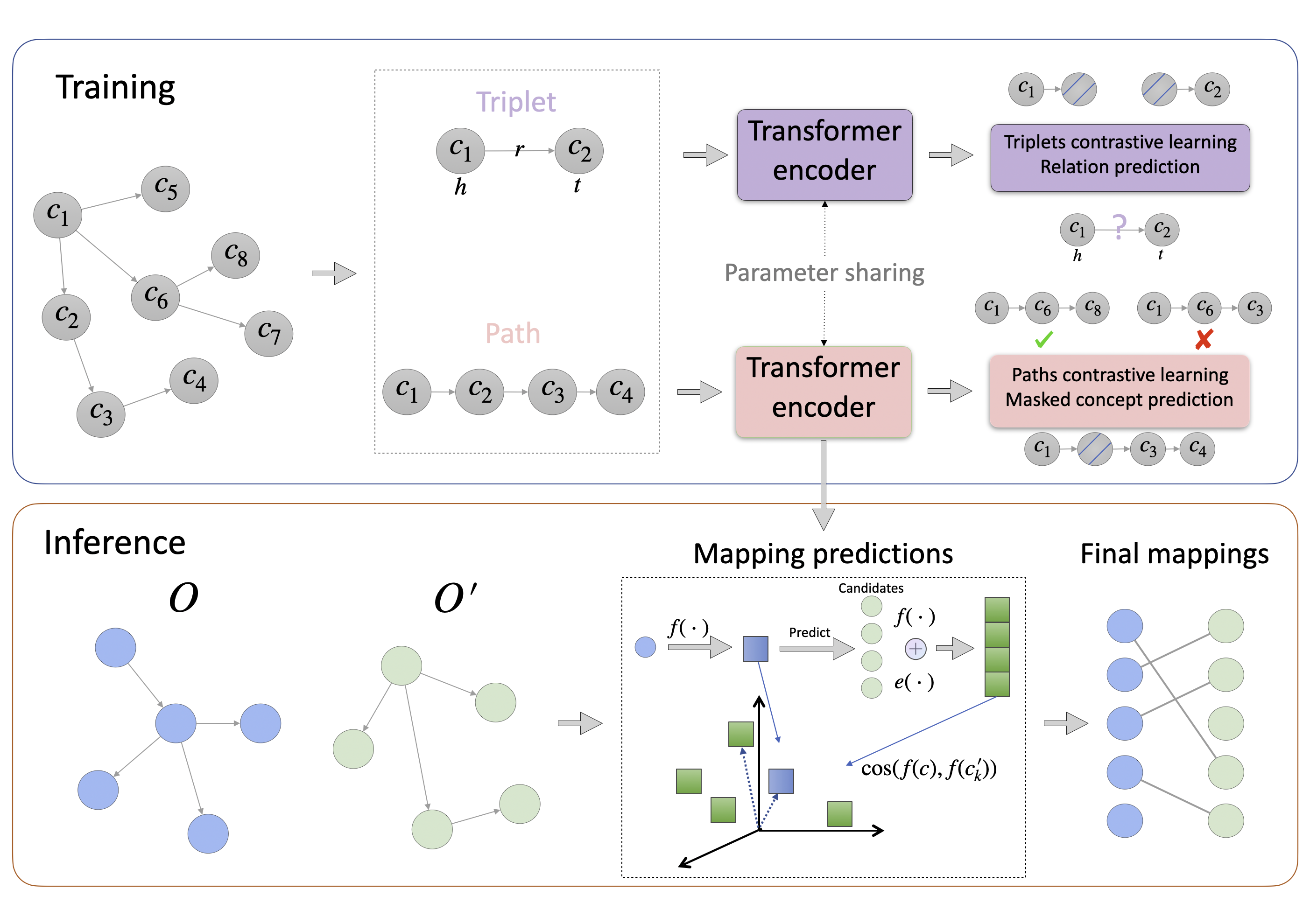}

\caption{\label{fig:framework}%
The overall framework of LaKERMap. In training, inputs are the ontologies. These ontologies are pre-processed for the corpus in triplets and paths. LaKERMap learns language and structure representations using transformers on multiple training objectives. During inference, inputs are source and target ontolgoies. It predicts candidates from the learned model, and obtains final mappings by calculating the cosine similarity. }
\vspace{-5mm}
\end{figure*}

{\bf LaKERMap Architecture.} The architecture of LaKERMap is illustrated in Figure \ref{fig:framework}. Initially, we construct triplet and path sets for the training corpus. LaKERMap incorporates two transformer encoders trained on different tasks, with all encoders sharing parameters. During inference, we extract the embeddings of a given concept $c$ from the learned encoders and generate target concept candidate sets as $(c_1',\ldots,c_k')$. Then, we compute the cosine similarity $s_i$ between $c$ and each $c_i', i\in k$, and select the final alignment based on similarity scores.

\subsection{Triplet Encoding} \label{sec:triplet}

Triplet-level representation learning considers a batch of triplets as inputs and estimates the joint distribution of elements in the triplets. We first construct the triplet set for each $c\in C$, and identify the relation set $R$ (e.g., subclass of, disjoint with, synonym, etc.). Then, we iterate over all the relations in KGs. Specifically, for each relation $r_i$, we aim to find the connected head and tail concepts to create a triplet $(h, r_i, t)$. All the triplets are subsequently passed to the transformer encoders to learn meaningful representations.


To incorporate contextual and local structural information in the triplets, we adopt multiple strategies for different relations to generate $n$ negative triplets for each positive pair. Notably, we mainly focus on the relation of `subclass of', such as (angiosarcoma, subclass of, sarcoma) and `synonym', such as (angiosarcoma, has synonym, hemangiosarcoma). To generate negative triplets for the `subclass of' relation, given a head concept $h$, we randomly sample $n$ tail concepts from those not in the subclass set. Differently, for the `synonym' relation, we first randomly sample $m$ tail concepts, which do not have overlapping tokens with the given head concept, from the subclass set.

{\bf Triplet contrastive learning.}
The mask prediction is core technique in PLMs, which is beneficial for contextualized representations and encouraging bidirectional context learning. Thus, we consider to randomly mask concepts in the triplets and predict them. To be specific, given a set of input $T = \{(h_i,r_i,t_i)\}_{i=1} ^ N$, we encode the distribution when one of the concepts is masked, i.g., $t_i$, written as:

\begin{equation}
    -E_{(h,r)\sim T} \log (P(f(t)\vert f(h),f(r)) 
\end{equation}
where $f(\cdot)$ represents the feature extraction module.

To improve robustness and performance on downstream tasks, such as ontology alignment, we also propose to learn from the contrastive representations of masked triplets. It is beneficial to maximize the similarity of matched pairs while minimizing the similarity of unmatched pairs \cite{pan2022contrastive}. For instance, the positive pair is $((h_i,r_i,[mask]),(([mask],[mask],t_i))$. We derive contrastive learning as a classification task and define Concept-Concept cross-entropy loss as:

\begin{equation} \label{eq:c2c}
    \mathcal{L}_{C2C} = - \sum_i^N \log (\frac{\exp(cos(f(h_i,r_i,[mask]), f([mask],[mask],t_i))/\tau)}{\sum_j\exp(cos(f(h_i,r_i,[mask]), f([mask],[mask],t_j))/\tau)})
\end{equation}

where $\tau$ is a scalar temperature hyperparameter.

{\bf Relation classification.} We also encode the distribution of relations when the relations are masked as [mask] token, written as:
\begin{equation}
    -E_{(h,r)\sim T} \log (P(f(r)\vert f(h),f(t))).
\end{equation}

There are various relation types connecting concepts. Given head and tail concepts $(h_i, [mask], t_i)$, we consider the relation prediction as a multi-class classification task. We extract relation representions through an auxiliary two-layer MLP and define Concept-Relation loss as:
\begin{equation}
    \mathcal{L}_{C2R} = -\sum_i^N \sum_{r\in R} y_i \cdot log(MLP(f(h_i,[mask],t_i)))
\end{equation}
where $y_i = 1$ is the true label of the i-th relation. Note that, we generate negative pairs in different ways according to the relation types.


\subsection{Path Encoding} \label{sec:path}
Path-level representation learning aims at capturing global interactions between the concepts. In particular, there are many paths from the root to the given concepts, some of which have significant long lengths. The hierarchical subclass paths are extracted from the graph structure of the ontologies. Additionally, we incorporate contextual information into the paths. 

For example, given triplets $(c_1, \text{subclass of}, c_2)$ and $(c_2, \text{synonym}, c_3)$, the path is $(c_1, \text{subclass of}, c_2, \text{synonym}, c_3)$. To generate negative paths, we randomly replace a concept for the given paths  with lengths less than 5, such as $(c_1, \text{subclass of}, c_5, \text{synonym}, c_3)$. In contrast, for paths with greater length, $20 \%$ concepts will be replaced randomly.


{\bf Path contrastive learning.} The goal is to differentiate between positive and negative paths by minimizing the contrastive loss. This training objective helps the model learn to identify meaningful paths and capture the global contextualized structural information among the ontologies. Given a set of paths $D = \{(c_i^1,c_i^2,\ldots, c_i^l)\}_{i=1}^Q $, we consider it as a binary classification task to predict the given paths are positive or negative. The CPath loss is defined as:
\begin{equation}
    \mathcal{L}_{cpath} = - \sum_{i=1}^{Q} \sum_{d_i\in D} y_i \cdot log(\sigma( f(d_i)))
\end{equation}

where $d_i$ is the $i$-th path $\in D$, $y_i$ is the label, $\sigma(\cdot)$ is the sigmoid function.

{\bf Masked concept prediction.} The advantages of transformer architecture provide the abilities to capture global structural information in the long sequences. Similar to the masked concept prediction in triplets, we randomly mask a concept in the path. Specifically, given a set of positive paths as $H =\{(c_i^1,c_i^2,\ldots, c_i^l)\}_{i=1}^M$, we feed the masked path such as as $(c_i^1,[mask],\ldots, c_i^l)$ to the transformer encoder.  The path encoding task is to predict the mask concept in the path, and we derive the distribution of masked concepts as:
\begin{equation}
    -E_{(c_i^1,\ldots,c_i^k)\sim H} \log (P(f(c_i^{mask})\vert f(c_i^1),\ldots, f(c_i^l)) 
\end{equation}
The MPath loss is defined as:
\begin{equation}
    \mathcal{L}_{mpath} = - \sum_{i=1}^{M} \sum_{c_i\in C} y_i^j \cdot log(softmax(f(c_i^j)))
\end{equation}
where $y_i^j$ is the label of masked concepts in the $i$-th path, $j$ is the masked position.

Finally, the overall training loss is as:
\begin{equation}
    \mathcal{L} = \mathcal{L}_{C2C} + \mathcal{L}_{C2R} + \mathcal{L}_{cpath} + \mathcal{L}_{mpath}
\end{equation}

\subsection{Mappings inference}
During the inference, the goal is to find an aligned concept from the target ontology for the given concept from the source ontology. Our model is self-supervised training on all ontolgoies in zero-shot settings. Thus, we consider the mapping problem as masked concept prediction. Specifically, given the concept $c_i$ from source ontology $O$, the predictions of concept candidate sets are derived as:
\begin{equation}
    \Omega(C) = softmax(f(c,c_{mask}))
\end{equation}
where $\Omega(C)=(c_1,\ldots,c_k)$ is the top-$k$ concepts based on probabilities. Then, we search each concept of the candidate sets in the target ontology $O'$.  

Inspired by filtering strategies in many OM systems \cite{wang2022amd,hertling2022ATBox}, we select final alignments by using a translation-based model, TransE \cite{bordes2013translating}, to encode relation information. To avoid increasing model complexity, we pre-process TransE embeddings and regard them as the relation regularizations. Then, we calculate similarity scores of the given concept by concatenating TransE embeddings to the outputs of the transformers. Suppose found $\Omega(C')=( c'_1,\ldots, c'_j) \in O',j \leq k$, we compute the cosine similarity for the mapping score as:
\begin{equation}
    S =(s_1,\ldots,s_j), 
\end{equation}
\begin{equation}
    s_j = cos(concat(f(c_i),e(c_i),concat(f(c'_j),e(c'_j))))
\end{equation}
where $e(\cdot)$ is the embedding from TransE.
Finally, the final mapping is selected by the top scored candidate as follows:

\begin{equation}
    M = (c,c',\equiv,argmax(S(c,c'))
\end{equation}
\section{Experiments}
We mainly conduct various experiments of LaKERMap on equivalence matching, and use Bio-ML track \footnote{\url {https://www.cs.ox.ac.uk/isg/projects/ConCur/oaei/2022/}}, developed by OAEI 2022, as the showcases. Note that, our method is  generalized and extendable to any tracks or datasets. The yearly results of OAEI illustrate the performances of state-of-the-art ontology matching systems. In this section, we introduce the implementation details of our model, its performance and analysis on different datasets, and ablation studies of different settings and training objectives.


\subsection{Implementation details}
{\bf Dataset.} We train on all ontologies without any reference mappings in Mondo and UMLS from Bio-ML track. The statistical details of the ontologies are shown in \cite{he2022machine}. During inference, it takes two particular ontologies as input files from the track. To evaluate, we use the subsets of equivlance matching task and unsupervised settings which are described in the track.

{\bf Ontology pre-processing.} Ontologies are usually formatted as owl or rdf, but the inputs of our model require the format of text tokens. Firstly, we extract meta-information from ontologies using owlready2 \footnote{\url{https://github.com/pwin/owlready2}}, including ID, labels, resources, and descriptions of concepts. The structural relations are extracted from subClass and disjointWith. Note that since the ontologies in the tasks were developed by different organizations, we process the ontology parsing from different tags, such as \textit{rdf:ID="isPartOf"}  and \textit{rdf:resource ="UNDEFINED\_part\_of"}. We construct sets of triplets and paths following the descriptions in sections 2.2 and 2.3. The details of our training set are presented in Table \ref{tab:stat}.
\begin{table}[]
\centering
\caption{The statistical details of pre-processed triplets and paths with instr-ontologies. We use them to construct the training corpus.} \label{tab:stat}
\begin{tabular}{c|c|c|cc|cc}
\hline
\multirow{2}{*}{Source} & \multirow{2}{*}{Ontology} & \multirow{2}{*}{\#Concepts} & \multicolumn{2}{c|}{\#Triplets} & \multirow{2}{*}{\#Paths} & \multirow{2}{*}{Avg. length} \\ \cline{4-5}
                        &                           &                             & subclass       & synonym        &                          &                              \\ \hline
\multirow{4}{*}{Mondo}  & DOID                      & 8,848                       & 28,113         & 131,964        & 19,566                   & 7.11                         \\
                        & NCIT                      & 6,835                       & 16,338         & 517,887        & 9,402                    & 3.91                         \\
                        & OMIM                      & 9,642                       & 12,966         & 502,359        & -                        & -                            \\
                        & ORDO                      & 8,838                       & 7,596          & 223,731        & 3,636                    & 4.47                         \\ \hline
\multirow{3}{*}{UMLS}   & FMA                       & 64,726                      & 206,775        & 2,412,222      & 63,867                   & 4.23                         \\
                        & NCIT                      & 29,206                      & 89,709         & 1,739,817      & 57,378                   & 3.76                         \\
                        & SNOMED                    & 51,498                      & 108,729        & 2,339,262      & 48,702                   & 4.48                         \\ \hline
                        
\end{tabular}
\vspace{-3mm}
\end{table}

{\bf Training settings.} In practice, we fine-tune BioBERT \cite{alsentzer-etal-2019-publicly} on our proposed training objectives. The number of fine-tuning epochs is 5 with batch size of 32 and learning rate of 5e-5. The other hypermater settings are as: $\tau$ in Eq \ref{eq:c2c} is 2; the positive-negative sample in both triplets and paths is 1:2; the number of candidate concepts in inference is 5. We train on 8 NVIDIA RTX 2080Ti GPUs, and infer on 1 NVIDIA RTX 2080Ti GPU. To compare the performance with other OM systems in OAEI, we adopt evaluation of DeepOnto \footnote{\url{https://github.com/KRR-Oxford/DeepOnto}}to avoid errors and bias during the evaluation process.

{\bf Evaluation Metrics.}
Following OAEI evaluation metrics, we evaluate our models in term of Precision, Recall and F1-score as follows:

\begin{equation}
    P = \frac{M \cap M_{ref}}{M}, R = \frac{M \cap M_{ref}}{M_{ref}}, F =  \frac{2 P  R}{P + R}
\end{equation}

where $P$ denotes Precision, $R$ denotes Recall, $F$ denotes F-score, $M$ is the output mappings of LaKERMap ,and $M_{ref}$ refers to the reference mappings which are annotated and verified by OAEI. Note that, we extract mapping pairs with equal relation from the references, and perform one-to-one mapping.
\begin{table}[]
\centering
\caption{Comparison to states-of-art OM systems in terms of precision, recall and F-score. \textbf{Best} scores and \underline{second-best} scores are highlighted. The sixth to twelfth rows are our model results with different training objectives. LaKERMap is training on all proposed objectives.} \label{tab:result1}
\begin{tabular}{c|ccc|ccc}
\hline
\multirow{2}{*}{Model} & \multicolumn{3}{c|}{OMIM-ORDO (Disease)}      & \multicolumn{3}{c}{NCIT-DOID (Disease)}       \\ \cline{2-7} 
                       & Precision     & Recall        & F-score       & Precision     & Recall        & F-score       \\ \hline
LogMap                 & 0.83          & 0.50          & 0.62          & {\ul 0.92}    & 0.67          & 0.77          \\
Matcha                 & 0.74          & 0.51          & 0.60          & 0.91          & 0.76          & 0.83          \\ \hline
BERTMap                & 0.73          & 0.57          & 0.64          & 0.91          & {\ul 0.83}    & {\ul 0.87}    \\
AMD                    & 0.66          & 0.57          & 0.61          & 0.89          & 0.77          & 0.82          \\
ATMatcher              & \textbf{0.94} & 0.25          & 0.39          & \textbf{0.96} & 0.60          & 0.74          \\ \hline
C2C                    & 0.71          & 0.45          & 0.55          & 0.89          & 0.78          & 0.83          \\
C2R                    & 0.78          & 0.49          & 0.60          & 0.87          & 0.75          & 0.81          \\
CPath                  & 0.75          & 0.36          & 0.49          & 0.86          & 0.73          & 0.79          \\
MPath                  & 0.68          & 0.42          & 0.52          & 0.82          & 0.69          & 0.75          \\
C2C + C2R              & 0.84          & {\ul 0.58}          & {\ul 0.69}    & 0.91          & 0.82          & 0.86          \\
CPath + MPath          & 0.75          &  0.46    & 0.57          & 0.88          & 0.77          & 0.82          \\
LaKERMap                    & {\ul 0.87}    & \textbf{0.61} & \textbf{0.72} & {\ul 0.92}    & \textbf{0.84} & \textbf{0.88} \\ \hline
\end{tabular}
\vspace{-3mm}
\end{table}

\subsection{Results and analysis}
We evaluate the performance of our proposed model in terms of evaluation metrics and runtime of mapping generations. In this section, we discuss the baselines we use, and the main results and analysis in different datasets.


{\bf Baselines.} The baselines and their results \footnote{\url{https://www.cs.ox.ac.uk/isg/projects/ConCur/oaei/2022}} are chosen from the participants in Bio-ML track in OAEI 2022. We mainly select Ontology Matching (OM) systems from two categories, including traditional OM matching with lexical and structure information (LogMap \cite{jimenez2022logmap} and Matcha \cite{faria2022matcha}), and machine/deep learning methods with pre-trained large language model and knowledge graph embbeding (BERTMap \cite{he2022machine}, AMD \cite{wang2022amd}, ATMatcher \cite{hertling2022ATBox}.) Note that, Matcha applies the same matching algorithms with AML \cite{faria2013agreementmakerlight} but with different ontology processing.

{\bf Main results and analysis.} The mapping results are shown in Tables \ref{tab:result1} and \ref{tab:result2}. Other model results are taken from OAEI 2022. We also provide the results of training on different combinations of triplet and path objectives. The analysis will be explained in the ablation studies section \ref{sec:ablation}. We can summarize that LaKERMap achieves the best or second-best performance compared with these state-of-the-art baselines. It outperforms the best baseline in terms of recall and F-score by $4 \%$ and $8 \%$, respectively. Although ATMatcher has the best precision scores, their performance on recall is not good. However, LaKERMap achieves the second-best scores in precision and surpasses other baselines by $4 \% - 11 \%$ in the OMIM-ORDO dataset.

Compared with the OM models that consider structural information, such as ATMatcher and AMD, LaKERMap is able to predict more correct mappings with contextualized structural representations on triplet-level and path-level. ATMatcher and AMD utilize limited structural contexts to filter mappings, which prevents them from achieving further improvements. BERTMap has shown the power of large language models compared to the lexical matching methods by LogMap and Matcha. However, BERTMap mainly focuses on learning contextual representations and repairing mappings with ontology structure, which neglects the sufficient structural information during training. Hence, our proposed model learns contextual and structural information seamlessly, and the integration of local and global structures can further improve the performance on mapping predictions. In summary, LaKERMap learns more generalized representations for the concepts and predicts high-quality mappings.


\begin{table}[]
\centering
\caption{Comparison to AMD and BERTMap in terms of runtime and evaluation metrics. \textbf{Best} scores are highlighted. In runtime, `s' denotes second, and `h' denotes hours.  } \label{tab:result2}
\begin{tabular}{c|cccc|cccc}
\hline
\multirow{2}{*}{Model} & \multicolumn{4}{c|}{SNOMED-FMA (Body)}                        & \multicolumn{4}{c}{SNOMED-NCIT (Pharm)}                      \\ \cline{2-9} 
                       & Runtime       & Precision     & Recall        & F-score       & Runtime      & Precision     & Recall        & F-score       \\ \hline
AMD                    & 1656(s)       & 0.89          & 0.70          & 0.79          & 452(s)       & 0.96          & 0.75          & 0.84          \\
BERTMAP                & $\sim$22(h) & \textbf{1.00} & 0.64          & 0.77          & $\sim$8(h) & \textbf{0.97} & 0.61          & 0.75          \\ \hline
LaKERMap                   & 321(s)        & 0.96          & \textbf{0.72} & \textbf{0.82} & 58(s)        & 0.94          & \textbf{0.78} & \textbf{0.85} \\ \hline
\end{tabular}
\end{table}

{\bf Efficiency analysis.} The runtime latency of an Ontology Matching (OM) system is another crucial evaluation metric. To compare the runtime of the mapping predictions, we reproduce AMD \footnote{\url{https://github.com/ellenzhuwang/AMD-v2}} and BERTMap \footnote{\url{https://github.com/KRR-Oxford/DeepOnto}} with the same computation resources as LaKERMap. As BERTMap adopts fine-tuning and mapping repair techniques during the predictions, it requires hours to generate the final alignments. In fact, it is impractical to match large scale ontologies. Compared with AMD, our model achieves significant improvements in terms of runtime and mapping quality, which results are shown in Table \ref{tab:result2}.

\subsection{Ablation studies} \label{sec:ablation}
We further evaluate the effectiveness of the different proposed encoding methods described in sections \ref{sec:triplet} and \ref{sec:path}. Specifically, we show the results of training on each objective, combining objectives on triplet-level and path-level separately. Moreover, we conduct a variety of experiments on the factors which impact the representation learning during training and mapping predictions in the inference. These include the ratio of positive-negative samples for corpus construction, the number of masked concepts in the path encoding, and the number of top-scored candidates in the mapping predictions.
\begin{figure}
\centering
     \begin{subfigure}[b]{0.32\textwidth}
        \centering
         \includegraphics[width=1.25\textwidth]{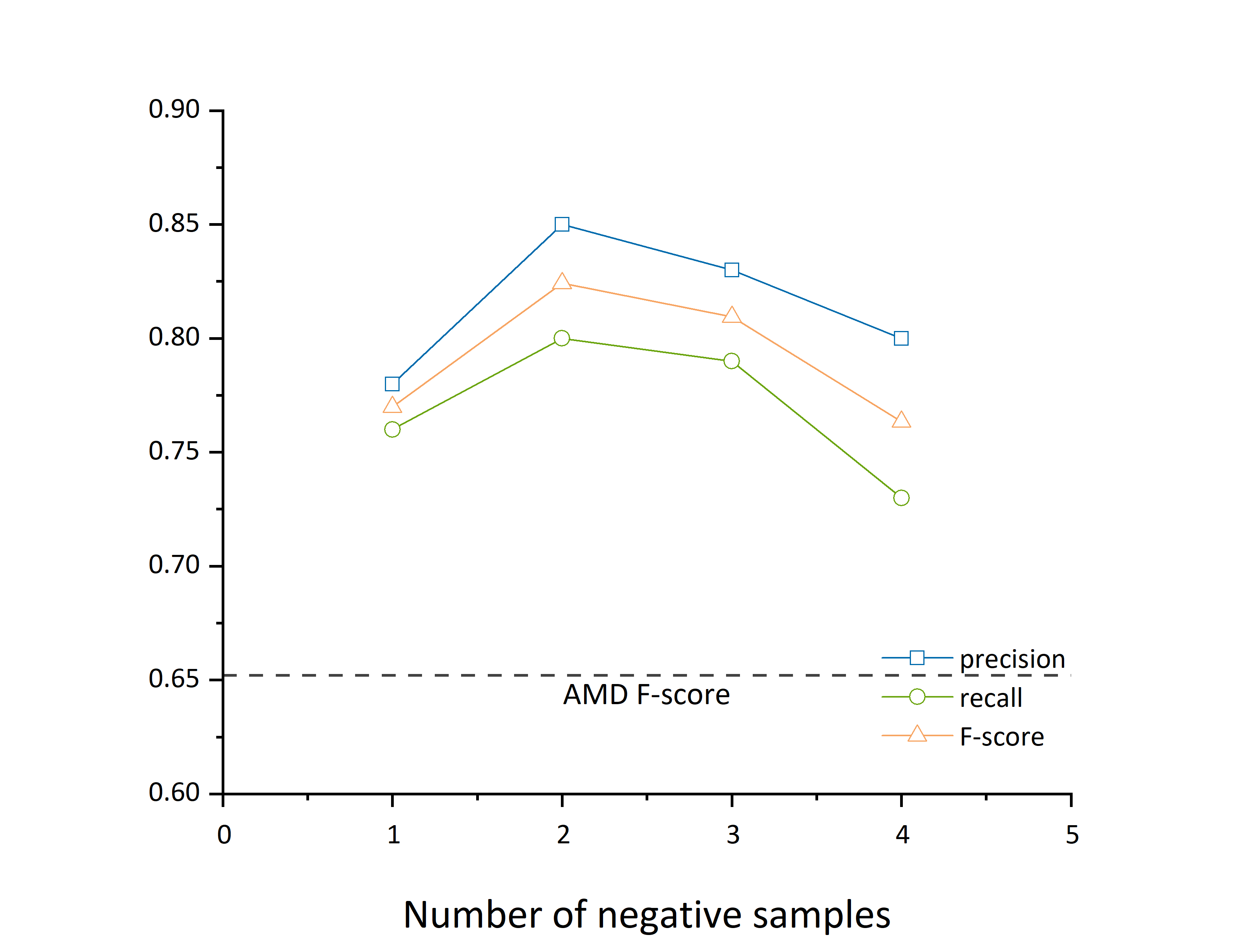}
         \caption{Results on different number of negative samples for each positive sample.}
         \label{fig:sample}
     \end{subfigure}
     \hfill
     \begin{subfigure}[b]{0.32\textwidth}
        \centering
         \includegraphics[width=1.25\textwidth]{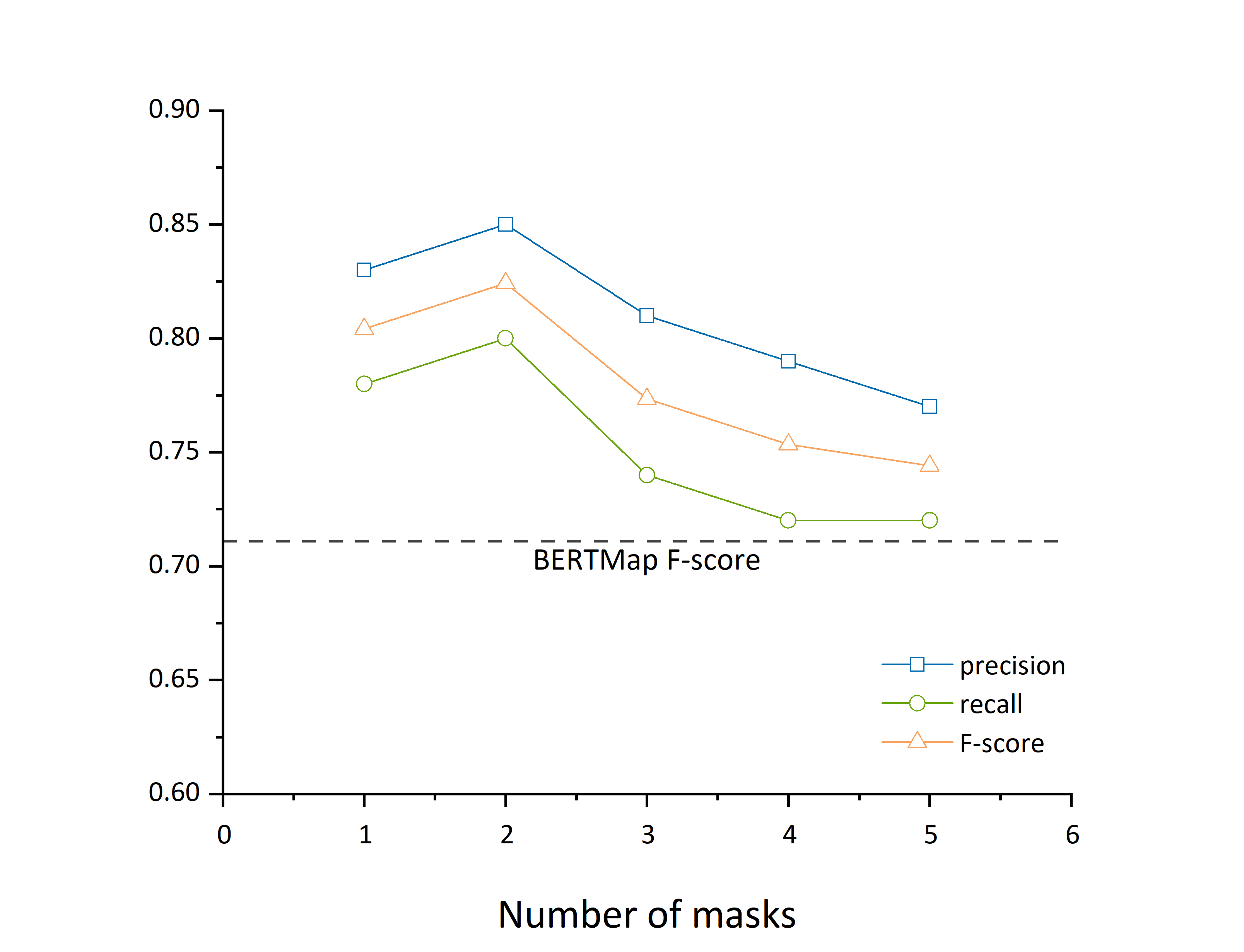}
         \caption{Results on different number of masks in masked path prediction.}
         \label{fig:mask}
     \end{subfigure}
     \hfill
     \begin{subfigure}[b]{0.32\textwidth}
        \centering
         \includegraphics[width=1.25\textwidth]{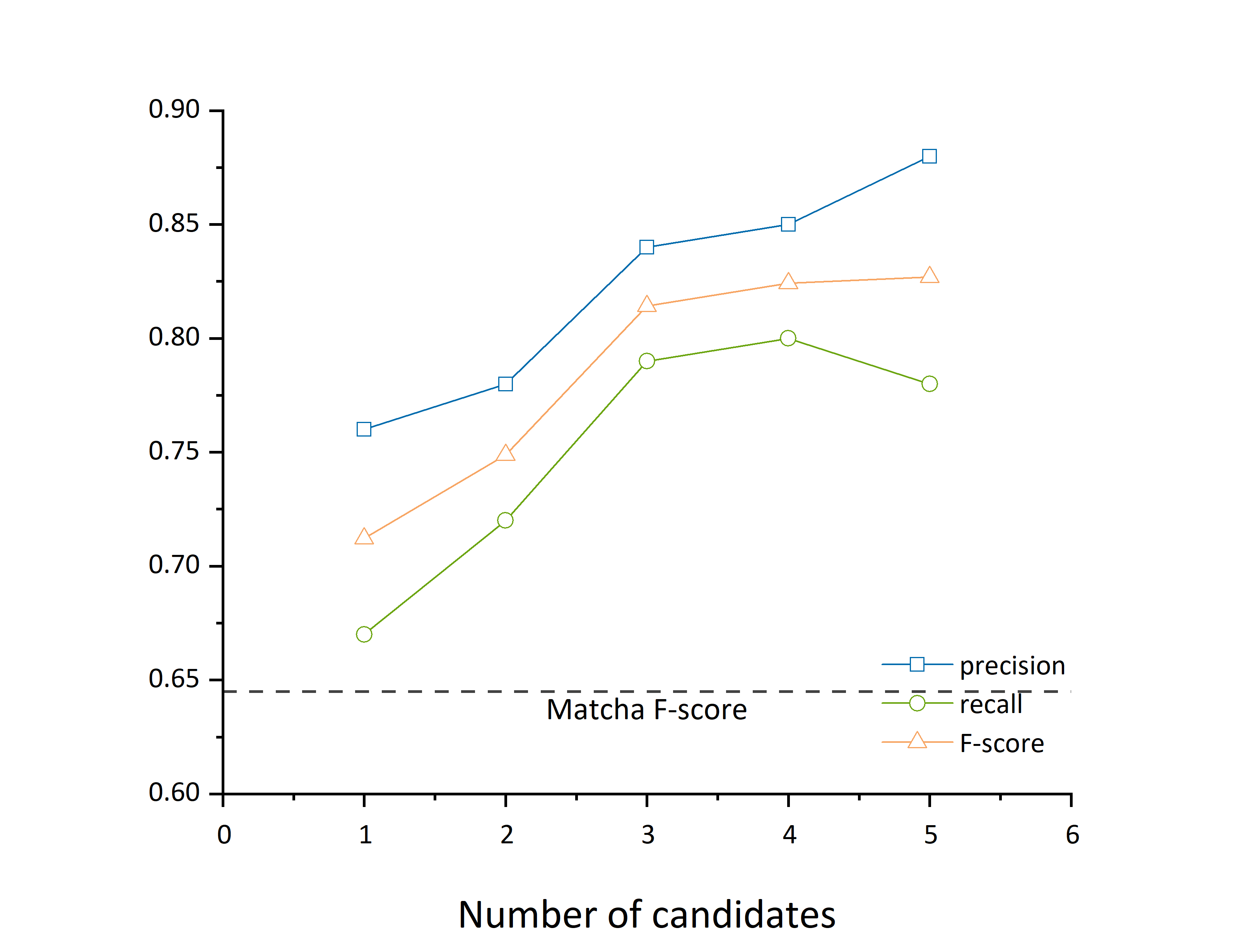}
         \caption{Results on different number of concept candidates in the inference.}
         \label{fig:candidate}
     \end{subfigure}
        \caption{Results of ablation studies on SNOMED-NCIT (Neoplas). The dash horizon lines are the F-scores from the baslines. }
        \label{fig:three graphs}
        \vspace{-5mm}
\end{figure}

\subsubsection{Which objective(s) from - Triplet or Path?}

We propose incorporating contextual and structural information with different training objectives. Evaluating each objective and their combinations in downstream tasks, specifically ontology matching in our case, proves to be beneficial. As shown in Table \ref{tab:result1}, we observe that triplet learning surpasses path learning in terms of F-score across both datasets, possibly due to the more significant semantic information captured by triplet encoding. When compared with solely using triplet encoding or path encoding, training on all objectives significantly increases the performance in term of all evaluation metrics. Therefore, encoding local and global interactions between concepts with contextual representations leads to the best performance in mapping predictions.

\subsubsection{Which ratio for masks and negative samples?} 

The quality of the corpus is pivotal in training deep learning models. Consequently, we conduct experiments on corpus construction, initially considering the number of negative samples for each positive sample. Negative samples in both triplet and path encoding can help distinguish between different samples, such as incorrect relations and paths. 

The results from varying the number of negative samples are shown in Fig \ref{fig:sample}. Positive-to-negative ratios of $1:2$ and $1:3$ outperforms the scenario with a ratio of $1:1$. Thus, negative samples are beneficial for improving generalization. However, an excess of negative samples can impair performance and robustness due to the bias introduced by imbalanced data.  


In path encoding, the model learns to understand and capture the semantic and syntactic relationships between concepts globally by predicting the masked concepts based on the surrounding concepts. The number of masked concepts impacts the model's understanding ability. Therefore, we illustrate the experiments on the average number of masks in Fig \ref{fig:mask}. It should be noted that the number of masks varies with the path length. Considering the average length of paths shown in Table \ref{tab:stat}, the performance declines with more than 2 masks. Since the model receives less information from the paths, more masks could potentially degrade the quality of the learned representations.

\subsubsection{How many candidates for inference?} 

During mapping inference, we predict concepts from the pre-trained model and employ knowledge graph embeddings to encode relations, which allows us to filter mappings from the candidate sets. As a result, the final alignments may differ from the model's outputs. For instance, the probability of $c'_j$ as $f(c'_j)$ might be the fifth highest in the candidate sets, but it may become the final alignment after considering $e(c'_j)$. 

As shown in Fig \ref{fig:candidate}, the recall score decreases as the number of candidates increases. Indeed, the final mappings are sensitive to the selections of the candidate sets in terms of precision and recall. In practice, we have observed that a candidate set size of 5 generally performs well.
\section{Related Work}
\subsection{Traditional Ontology Matching} 


Traditional ontology matching systems, such as AgreementMaker\cite{vldb2009demo} and AgreementMakerLight \cite{faria2013agreementmakerlight}, typically employ feature-based methods. As leading ontology matching systems, they deploy high flexibility and extensibility by relying on a combination of various approaches, including lexical matching, structure matching, and external knowledge resources. The LogMap family~\cite{jimenez2011logmap}, another prominent system in OAEI, extends mapping by adopting structural information and repairing mapping through logical reasoning. While these systems have been proven to be effective over the past decades, their matching processes rely heavily on outdated NLP techniques such as lexical matching, thereby neglecting the enrichment of contextual semantics.


Recent methods have begun to apply word embeddings. DeepAlignment~\cite{kolyvakis2018deepalignment} and SCBOW~\cite{kolyvakis2018biomedical} leverage representation learning for ontology matching problems. They use word embeddings such as Word2Vec \cite{mikolov2013efficient} to extract synonyms with joint structure encoding. ALOD2Vec~\cite{portisch2018alod2vec} utilizes large RDF data with hypernymy relations as external resources, deriving embeddings for concepts based on the RDF2Vec~\cite{ristoski2016rdf2vec} method. However, much like AML and LogMap, these models focus on feature engineering and do not sufficiently consider semantics.

\subsection{Machine/Deep learning models based Ontology Matching}  Recent works, such as MultiOM \cite{li2019multi}, perform the matching process using different embedding techniques from multiple views in graphs. In their work, they compare different knowledge graph embeddings evaluating on large biomedical ontologies. VeeAlign~\cite{iyer2020veealign} proposes a dual attention mechanism for fusing concepts and their neighbour weights. Log-ML~\cite{chen2021augmenting}, a machine learning extension built atop LogMap, trains a complex neural network to incorporate semantics. However, some of these models require seed or reference alignments. Moreover, they are not particularly efficient in mapping prediction, which typically requires high computational resources and several hours to generate the ontology alignments.

As an extension of AML, AMD~\cite{DBLP:conf/semweb/WangC21} adopts a pre-trained transformer model to generate mappings and filters using knowledge graph embedding methods. Neutel et al.\cite{neutel2021towards} extract embeddings from Sentence-BERT\cite{reimers2019sentence} for ontology matching. BERTMap~\cite{he2021bertmap} is a BERT-based system that fine-tunes BioBERT~\cite{alsentzer-etal-2019-publicly} on a semantic text corpus derived from ontologies. MELT~\cite{hertling2021matching} incorporates a transformer-based filter in their machine learning module and discusses performance based on different pre-trained language models and fine-tuning settings. However, these methods mainly focus on textual semantic representations, thereby neglecting important structural information in the ontologies.

\section{Conclusions}

In this paper, we propose capturing both contextual and structural information of the ontologies using distinct training objectives. LaKERMap is composed of two transformers, encoding triplets and paths respectively. We train LaKERMap in a self-supervised manner and infer mappings in a zero-shot setting. Specifically, we fuse and feed contextual information, such as label names, and structural information, such as relations between concepts, into the transformers. The training tasks encompass triplet contrastive learning, relation classification, path contrastive learning, and masked concept prediction. 

Through the benefits of self-supervised learning, our model learns generalized representations of the concepts and can generate mappings within seconds. By conducting extensive experiments on different datasets and in various settings, we demonstrate that LaKERMap surpasses state-of-the-art baseline ontology matching systems in terms of speed and accuracy.


{\bf Limitations and future work.} Due to computational resource limitations, we initialize and fine-tune BioBERT as our backbone encoder. Performance could potentially be improved by adopting larger and more complex language models, such as LLaMA\cite{touvron2023llama}. Furthermore, we believe that self-supervised training on a broader range of ontologies will enhance generalization and robustness. Specifically, we plan to incorporate additional ontology resources, such as BioPortal \cite{noy2009bioportal}, with the aim of providing a large-scale pre-trained model for ontology matching as part of our future work.


\bibliography{bibliography}

\end{document}